\pdfoutput=1

\documentclass[11pt]{article}

\usepackage{array}
\newcolumntype{P}[1]{>{\centering\arraybackslash}p{#1}}
\newcolumntype{M}[1]{>{\centering\arraybackslash}m{#1}}

\usepackage{graphicx}

\usepackage[]{acl}

\usepackage{times}
\usepackage{latexsym}

\usepackage[T1]{fontenc}

\usepackage[utf8]{inputenc}

\usepackage{microtype}
\usepackage{multirow}

\usepackage{color,soul}

\title{Error syntax aware augmentation of feedback comment generation dataset}

\author{Nikolay Babakov$^1$, Maria Lysyuk$^1$, Alexander Shvets$^3$, Lilya Kazakova$^2$, and \\ \textbf{ Alexander Panchenko}$^1$\\
$^1$Skolkovo Institute of Science and Technology \\
$^2$National Research University Higher School of Economics \\
$^3$Pompeu Fabra University \\
\href{mailto:alexander.shvets@upf.edu}{\{n.babakov, a.panchenko,  maria.lysuk\}@skol.tech,   alexander.shvets@upf.edu} \\ 
}

\begin{document}
\maketitle
\begin{abstract}

This paper presents a solution to the GenChal 2022 shared task dedicated to feedback comment generation for writing learning. In terms of this task given a text with an error and a span of the error, a system generates an explanatory note that helps the writer (language learner) to improve their writing skills. Our solution is based on fine-tuning the T5 model on the initial dataset augmented according to syntactical dependencies of the words located within indicated error span. The solution of our team ``nigula'' obtained second place according to manual evaluation by the organizers.

\end{abstract}

\section{Introduction}

Feedback comment generation for language learners is the task of generating an explanatory note that helps the writers (language learners) improve their writing skills for text with error and a span of this error.
In terms of GenChall2022\footnote{\url{https://fcg.sharedtask.org/task/}}, the target language is English, but this task applies to any language. Moreover, this task is mostly designed not for such cases as simple typos or misspellings (something which can be easily detected by standard grammar-error detection systems) but for erroneous, unnatural, or problematic words. See \tablename~\ref{tab:table_samples}  for examples of such types of errors and corresponding comments.

Detecting the aforementioned misuse of words is not enough to prevent the same error in the future. It is important to provide some explanation and reference to the error-specific rule and give a direct hint on which correct word should be used in this particular case.  

While any language follows certain rules which can be encoded manually to obtain the solution relying on rules and templates, in our work we try to use the benefit of the availability of parallel data and explore the limits of modern seq2seq models with the minimal number of rules and manual labor involved.

The contributions of our work are as follows:
\begin{itemize}
    \item We present our solution based on tuning the T5-large model on the dataset augmented in the special error syntax-based approach.
    \item We opensource the model on Huggingface Model Hub\footnote{\url{https://huggingface.co/SkolkovoInstitute/GenChal_2022_nigula}} and code for experiments on GitHub\footnote{\url{https://github.com/skoltech-nlp/feedback_generation_nigula}}.
\end{itemize}

\section{Related works}

The first attempts to provide feedback on a particular error type were based on rules~\cite{nagata-etal-2014-correcting}. In~\citet{morgado-da-costa-etal-2020-automated} authors used the English Resource Grammar parser to analyze the learner's sentence. If the parser fails to process a sentence, this sentence is supposed to have an error, and, so-called, mal-rules are used to detect the particular type of error. If the mal-rule works, the user is provided with a mal-rule-specific comment.


\begin{table*}[]
\centering
\begin{tabular}{M{0.4\linewidth}M{0.5\linewidth}}
\hline
\textbf{Learner's sentence} & \textbf{Golden feedback / Our system's output} \\ \hline
\multirow{2}{*}{\shortstack{Maybe at holiday and have free time \\ then I can \textcolor{red}{to} have part-time job .}} & <Verbs> that follow an <auxiliary verb> are used in their <infinitive form> instead of a <to infinitive>. \\ \cline{2-2} 
 & <verbs> that come right after an <auxiliary verb> are used in their <infinitive form>. \\ \hline
\multirow{2}{*}{\shortstack{\textcolor{red}{Become} college student \\ requires a lot of money .}} & A <verb phrase> needs to be converted into into a <noun phrase> in the form of a <to infinitive> or a <gerund> to be used as the <subject>. \\ \cline{2-2} 
 & A <verb phrase> needs to be converted into into a <noun phrase> in the form of a <to infinitive> or a <gerund> to be used as the <subject>. \\ \hline
\multirow{2}{*}{\shortstack{They might face \textcolor{red}{with} the danger \\ of exploring of the battery or the  \\  problems of the electronic .}} & Since the <verb> <<face>> is a <transitive verb> and its <direct object> indicates the confronted object, it does not require a <preposition>. \\ \cline{2-2} 
 & Since the <verb> <<face>> is a <transitive verb>, the <object> does not require a <preposition>. \\ \hline
\multirow{2}{*}{\shortstack{There are many advantages \\ \textcolor{red}{to have} a part-time job .}} & Use <preposition + gerund> instead of a <to-infinitive> to describe the ``advantage''. Look up the use of the <noun> <<advantage>> in a dictionary. \\ \cline{2-2} 
 & Use the structure <preposition+gerund> instead of a <to-infinitive> with the <noun> <<advantage>>. \\ \hline
\end{tabular}
\caption{\textbf{Input and output (golden and system).} Example of feedback comments from the competition test set and generated without system. The word with an error is highlighted in red. < > signs indicate grammar terms, << >> mean reference to the word in the learner text.}
\label{tab:table_samples}
\end{table*}

The steps towards the usage of more modern approaches, such as neural networks, have also been performed in this task. In~\citet{andersen-etal-2013-developing}, the authors use a combination of basic machine learning approaches to detect errors and rules to provide feedback on some common types of errors. \citet{nagata-2019-toward} showed that a neural retrieval-based method can be effective in preposition feedback comment generation. \citet{lai-chang-2019-tellmewhy} proposed a method that provides context-sensitive explanations using grammatical error correction and templates. \citet{gkatzia-etal-2014-comparing} suggested methods for automatically choosing feedback templates based on learning history. In~\citet{https://doi.org/10.48550/arxiv.2203.07085}, the sentence with an error is first corrected with the grammar error correction system and then the K-nearest neighbors algorithm is used to provide the learner with the pair of an incorrect and corrected sentence which contains a similar kind of error. In~\citet{getman2021automated}, the authors use unusual n-grams, out-of-vocabulary words, and several pre-trained models to find an error in the learner's text. This system does not provide text feedback in natural language, but it generates a structured report of found errors in the text. 

The useful subtask of the feedback comment generation is grammar error classification. The information on the particular type of error made in the text could be used either directly by creating a template comment to this error or by using the error type as an additional signal in training data. One example of such work is ~\citet{bryant-etal-2017-automatic} which automatically extracts the edits between parallel original and corrected sentences using a linguistically-enhanced alignment algorithm. In this paper, a rule-based framework that relies solely on dataset-agnostic information such as lemma and part-of-speech is developed as well. Beyond this, the paper of \citet{choshen-etal-2020-classifying} uses universal dependencies syntactic representation scheme.


The main limitation of using the most modern text-to-text models had been the non-availability of parallel datasets with errors and corresponding annotation. In~\citet{pilan-etal-2020-dataset}, a unique dataset where feedback comments on linking words were annotated was released. The dataset used in GenChal 2022 was collected in~\citet{nagata-2019-toward,nagata-etal-2020-creating} for the English language. The main types of errors in this dataset are misuse of prepositions and other writing items such as discourse and lexical choice. 

\section{Task description}
In this section, we introduce the formal definition of the task and the dataset provided for it.

\begin{figure*}[ht]
\centering
\includegraphics[scale=0.85]{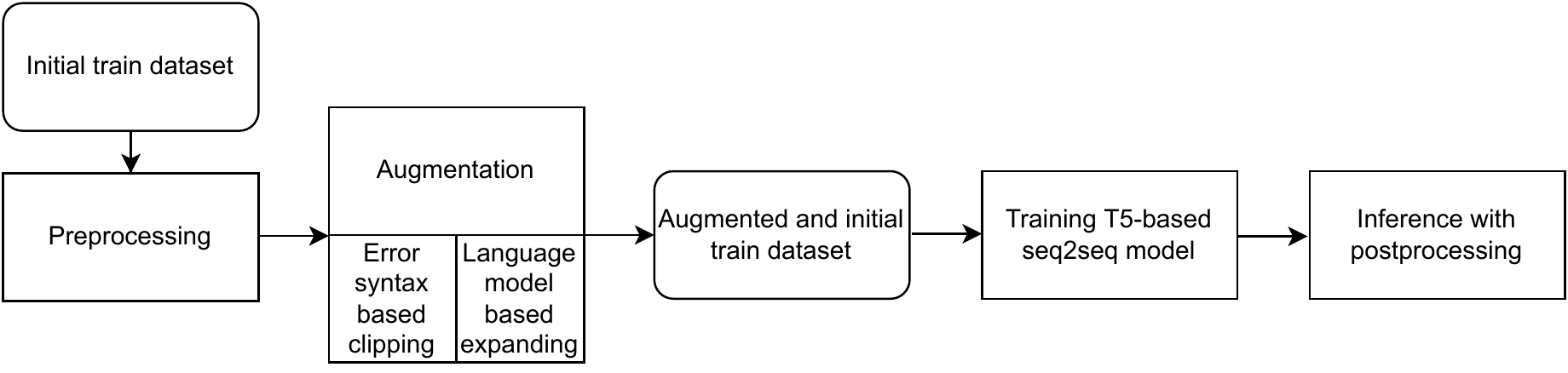}
\caption{\textbf{Method workflow.} Description of the main steps of our feedback generation approach. The initial dataset is first preprocessed, then it is augmented by clipping the learner sentences according to the syntax relations of the word within error spans and the clipped sentences are then expanded with a large language model. The initial and augmented versions of the dataset are then used to train the T5 model in seq2seq mode. After that, the trained model is used to generate the feedback comments on the test data and the final text is post-processed.}
\label{fig:pipeline}
\end{figure*}

\subsection{Task definition}
\label{sec:task_definition}

The system is provided with the text that by default contains an error. Moreover, the exact error span is provided as well. The output of the system should be the text which provides explanatory feedback on the error. If the system fails to generate the feedback it is supposed to return the <NO\_COMMENT> string.

\subsection{Dataset}
\label{sec:dataset}

The form of the data in the task dataset is as follows:

\begin{itemize}
    \item I agree it . \textbackslash t  3:10 \textbackslash t  \textless\phantom{-}\textless Agree\phantom{-}\textgreater\phantom{-}\textgreater\phantom{-} is an <intransitive verb> and thus it requires a <preposition> before its object.
\end{itemize}

where \textbackslash t stands for the tab character. If a sentence contains more than one error, it appears two or more times with different spans, so the input text always consists of only one sentence with one span. Also, the texts are pre-tokenized where tokens are separated by whitespace.

The feedback texts contain special symbols: <, >  for grammatical terms (e.g. <intransitive verb>, <noun>, etc.) and \textless\phantom{-}\textless, \phantom{-}\textgreater\phantom{-}\textgreater\phantom{-} for citations of the words from learners' sentence (e.g. <<agree>>).

The dataset consists of the train (4867 samples) and development (169 samples) sets which were provided at the beginning of the competition and the test set (214 samples) which was provided in the last week of the competition and had only text and error span information. 


\section{Our method description}

In this section, we introduce the main steps for training the model used for the final submission. These steps are also shown in Figure~\ref{fig:pipeline}.

\begin{table*}[h!]
\centering

\begin{tabular}{P{0.4\linewidth}P{0.4\linewidth}}
\hline
\textbf{Sentence} & \textbf{Comment} \\ \hline
They can help their father or mother about money that we must use in the university too . & \textless\phantom{-}\textless About \textgreater\phantom{-}\textgreater is not the appropriate \textless preposition\textgreater to be used when a \textless noun\textgreater follows the structure \textless help + someone \textgreater . Look up the use of the \textless verb\textgreater\phantom{-}\textless\phantom{-}\textless help \textgreater \phantom{-} \textgreater \phantom{-} in a dictionary to learn the appropriate \textless preposition\textgreater\phantom{-} to be used. \\ \hline
they can help their father or mother \phantom{-} \textless \phantom{-} \textless \phantom{-} about \phantom{-} \textgreater \phantom{-} \textgreater \phantom{-} money that we must use in the university too . & \textless \phantom{-} \textless \phantom{-} about \textgreater \phantom{-} \textgreater \phantom{-} is not the appropriate \textless\phantom{-}preposition\phantom{-}\textgreater\phantom{-} to be used when a \textless \phantom{-} noun \phantom{-}\textgreater \phantom{-}follows the structure \textless \phantom{-} help + someone \phantom{-}\textgreater \phantom{-}. look up the use of the \phantom{-} \textless \phantom{-} verb \phantom{-} \textgreater \phantom{-} \textless \phantom{-} \textless \phantom{-} help \textgreater \phantom{-} \textgreater \phantom{-} in a dictionary to learn the appropriate \phantom{-} \textless \phantom{-} preposition\phantom{-}\textgreater\phantom{-} to be used . \\ \hline

\end{tabular}
\caption{\textbf{Preprocessing.} Example of the preprocessing of learner's text and the corresponding feedback comment (the initial one is at the top, the preprocessed one is at the bottom).}
\label{tab:preprocessing}
\end{table*}

\subsection{Preprocessing}
\label{sec:preprocessing}

Even though the text in the dataset is pre-tokenized, the special symbols described in~\ref{sec:dataset} (<, >, <\phantom{-}< , >\phantom{-}>) can interfere with the tokenizer of a large pre-trained model. Thus, we lowercase the text and insert whitespace between words and the special symbols. Refer to \tablename~\ref{tab:preprocessing} to have a look at the example of an initial and preprocessed sample.

As mentioned in~\ref{sec:dataset}, one sentence can have multiple errors, but according to the task definition, the system is supposed to provide feedback only to one slot of the error. To explicitly point out the exact error span in the learner's sentence we put similar special symbols (\textless\phantom{-}\textless\phantom{-}, \phantom{-}\textgreater\phantom{-}\textgreater\phantom{-}) around the error span.

\subsection{Augmentation}
Even though the dataset has a limited amount of error types the variability of natural language yields an almost unlimited amount of situations in which each particular error can occur. 

Let's demonstrate it in the following example of learner's text - \textit{They can help their father or mother about money that we must use in the university too}. According to the span, the error is in the misuse of the preposition "about". To be more specific, the student has used an incorrect preposition after the "help + someone" construction. So, if we generate a new sentence that would be similar to the initial one by 'help someone about something' construction and would be different from other points of view, it will still correspond to the initial feedback and it could be applied to training the model in seq2seq mode as an additional training sample. 

\begin{figure*}[ht]
\centering
\includegraphics[scale=0.9]{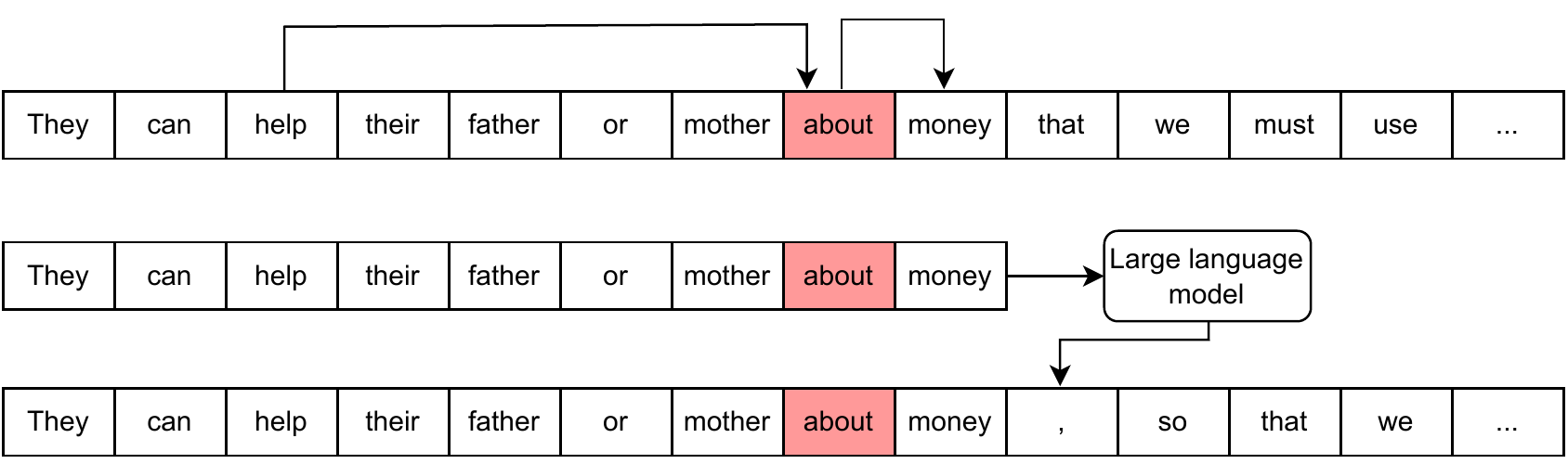}
\caption{\textbf{Augmentation workflow.} The sentence with an error passes automatic syntactical analysis. The sentence is cropped by the last word that is syntactically connected to any word within the error span or by the error span itself (if no syntactically dependent words are located after the span). The cropped sentence is used as a prompt for a large language model to generate an alternative phrase with a similar error. }
\label{fig:augment_pipeline}
\end{figure*}

Our approach to augmentation consists of two parts. First, we cut the initial sentence by the last word which is syntactically related to the words within an error span. Second, we use the remaining text as a prompt to the language model, so it generates an alternative end to the sentence with a given prefix. Refer to Figure~\ref{fig:augment_pipeline} for the principal scheme of the augmentation approach. More details can be found below in Sections~\ref{sec:clipping} and~\ref{sec:generation}.

\subsubsection{Learner sentence clipping}
\label{sec:clipping}

We use the spacy\footnote{\url{https://spacy.io/models/en##en_core_web_md}} package to perform syntactic parsing of the learner text. In the case of the sentence from our example, the error word "about" is syntactically connected to the verb "help" and the noun "money". We assume that these words form the skeleton of the error in this particular sentence. Thus, we crop everything after the last connected word 'money' and the cropped sentence becomes \textit{they can help their father or mother about money}. If the syntactically connected words are located before the error span we crop the sentence by the error span.

\subsubsection{Text generation}
\label{sec:generation}

The cropped version of the text can be used as a prompt to infer an entire sentence with a large pre-trained language model. We use EleutherAI/gpt-neo-1.3B~\cite{gao2020pile,gpt-neo}\footnote{\url{https://huggingface.co/EleutherAI/gpt-neo-1.3B}} to generate new texts. 

This approach allows us to get a new sample consisting of the new sentence (which consists of the prefix similar to the initial one and the extension generated by the language model) and the feedback similar to the initial sample.

Here are some examples of the sentences generated by the model from the prompt \textit{they can help their father or mother about money}:

\begin{itemize}
    \item they can help their father or mother about money, so that we can be independent. We have to work hard to earn our bread. 
    \item they can help their father or mother about money." "Well, if I do that I'll have to buy clothes, and pay my own expense." "The girl has
    \item they can help their father or mother about money, they can help the mother, who can give us some medicine? We are able to keep the household alive from the old and the sick with
\end{itemize}

Some examples in the original dataset have similar feedback comments. If the learner's sentence is related to the group of samples which have ten or more similar feedback sentences, we assume that there is enough information for training the final model and do not apply augmentation to it.

Thus, we apply the augmentation technique to more than 4700 samples from the initial dataset. Each sample is augmented with 8-10 sentences. The final versions of the augmented dataset consist of 43,174 samples. We use these samples as additional data for training the final model. 

\subsection{Model training}

The final solution is based on tuning T5 model~\cite{2020t5}\footnote{\url{https://huggingface.co/t5-large}}. The model's training input is the preprocessed sentences (see Section~\ref{sec:preprocessing}) and its target output is the corresponding feedback comments.  Our default training parameters are batch size 8, Adam optimizer, gradient clipping by 1, and learning rate 1e-5. During training, we evaluate the current version of the model by calculating the BLEU score~\cite{papineni2002bleu} on the validation dataset.

\begin{table}[]
\begin{tabular}{p{0.15\linewidth}p{0.4\linewidth}p{0.2\linewidth}}
\hline
\textbf{Exp. \#} & \textbf{Data}                  & \textbf{BLEU} \\ \hline
1             & Initial               & 0.64            \\ 
2             & Augmented             & 0.65            \\ 
3             & Initial\&augmented & 0.67            \\ \hline
\end{tabular}
\caption{ \textbf{Training steps.} Sequence of experiments conducted to train the final model. BLEU is shown for the validation dataset.}
\label{tab:experiments}
\end{table}

As shown in Table~\ref{tab:experiments} the main training steps are as follows:
\begin{itemize}
    \item We train the model on the initial dataset. The best validation BLEU score is 0.64
    \item We tune the best version of the model on the augmented dataset (just newly generated samples). The best validation BLEU score is 0.66
    \item We merge both datasets, decrease the learning rate to 1e-6 and tune it for 4000 steps. This increases the validation BLEU score up to 0.67
\end{itemize}

The final version of the model is open-sourced to HuggingFace ModelHub.

\subsection{Postprocessing}

During inference, the final model demonstrates unusual behavior in terms of placing the special marks. To be more specific, the model correctly generates the closing brackets (>, > >) but fails to generate opening brackets (<, < <). We apply some postprocessing to the generated text. For doing this, we collect all possible grammar term references within train and validation texts, find the longest possible term before the closing bracket, and put opening brackets before the found term.

\begin{table*}[ht!]
\begin{tabular}{P{0.25\linewidth}P{0.25\linewidth}P{0.25\linewidth}P{0.15\linewidth}}
\hline
\textbf{Preprocessed input} & \textbf{Reference comments} & \textbf{System output} & \textbf{Manual evaluation by orgs} \\ \hline
most \phantom{-}\textless \phantom{-}\textless\phantom{-} of \phantom{-}\textgreater \phantom{-}\textgreater\phantom{-} restaurant separate smoking are and non smoking are , but the smoking smell is remain this place . & Something defined follows \phantom{-}\textless{}\phantom{-}\textless{}\phantom{-}most of\phantom{-}\textgreater{}\phantom{-}\textgreater\phantom{-}{}. When referring to restaurants in general, use \phantom{-}\textless{}\phantom{-}\textless{}\phantom{-}most\phantom{-}\textgreater{}\phantom{-}\textgreater\phantom{-} as an \textless{}adjective\textgreater instead of a \textless{}noun\textgreater{}. & A group of something specific follows \phantom{-}\textless{}\phantom{-}\textless{}\phantom{-}most of\phantom{-}\textgreater{}\phantom{-}\textgreater\phantom{-}{}. Use \phantom{-}\textless{}\phantom{-}\textless{}\phantom{-}most\phantom{-}\textgreater{}\phantom{-}\textgreater\phantom{-} as an \textless{}adjective\textgreater rather than a \textless{}noun\textgreater when simply referring to something in general. & correct \\ \hline
most of restaurant separate smoking are and non smoking are , but the smoking smell is \phantom{-}\textless \phantom{-}\textless\phantom{-} remain \phantom{-}\textgreater \phantom{-}\textgreater\phantom{-} \phantom{-}\textless \phantom{-}\textless\phantom{-} this \phantom{-}\textgreater \phantom{-}\textgreater\phantom{-} place . & The \textless{}verb\textgreater \phantom{-}\textless{}\phantom{-}\textless{}\phantom{-}remain\phantom{-}\textgreater{}\phantom{-}\textgreater\phantom{-} is an \textless{}intransitive verb\textgreater and cannot take \textless{}direct objects\textgreater without a \textless{}preposition\textgreater of place. & When the \textless{}verb\textgreater \phantom{-}\textless{}\phantom{-}\textless{}\phantom{-}remain\phantom{-}\textgreater{}\phantom{-}\textgreater\phantom{-} is used as an \textless{}intransitive verb\textgreater{}, a \textless{}preposition\textgreater needs to precede the \textless{}object\textgreater{}. Look up the \textless{}verb\textgreater \phantom{-}\textless{}\phantom{-}\textless{}\phantom{-}remain\phantom{-}\textgreater{}\phantom{-}\textgreater\phantom{-} in the dictionary to find the appropriate \textless{}preposition\textgreater{}. & correct \\ \hline
part time job is a very important \phantom{-}\textless \phantom{-}\textless\phantom{-} of \phantom{-}\textgreater \phantom{-}\textgreater\phantom{-} people in present . & Use the \textless{}preposition\textgreater `for' to express ``to be important for people''. & It is not grammatically incorrect to use the \textless{}preposition\textgreater \phantom{-}\textless{}\phantom{-}\textless{}\phantom{-}of\phantom{-}\textgreater{}\phantom{-}\textgreater\phantom{-}{}, but using \phantom{-}\textless{}\phantom{-}\textless{}\phantom{-}of\phantom{-}\textgreater{}\phantom{-}\textgreater\phantom{-} emphasizes the part in one's opinion''. It is more natural to use the \textless{}preposition\textgreater for'to express to be important for...''. & incorrect \\ \hline
part time job is a very important of people \phantom{-}\textless \phantom{-}\textless\phantom{-} in \phantom{-}\textgreater \phantom{-}\textgreater\phantom{-} present . & The \textless{}preposition\textgreater used in the  \textless{}idiom\textgreater with the \textless{}noun\textgreater \phantom{-}\textless{}\phantom{-}\textless{}\phantom{-}present\phantom{-}\textgreater{}\phantom{-}\textgreater\phantom{-} to express ``now'' is not \phantom{-}\textless{}\phantom{-}\textless{}\phantom{-}in\phantom{-}\textgreater{}\phantom{-}\textgreater\phantom{-}{}. & The \textless{}preposition\textgreater used in the \textless{}idiom\textgreater with the \textless{}noun\textgreater \phantom{-}\textless{}\phantom{-}\textless{}\phantom{-}present\phantom{-}\textgreater{}\phantom{-}\textgreater\phantom{-} to express for the moment'' or now'' is not \phantom{-}\textless{}\phantom{-}\textless{}\phantom{-}in\phantom{-}\textgreater{}\phantom{-}\textgreater\phantom{-}{}. & correct \\ \hline
\end{tabular}
\caption{Examples of system output to similar sentences with different error span}
\label{tab:similar_sentences_output}
\end{table*}

\subsection{Other experiments}

Except for the data augmentation approach, we also tried other hypotheses, which did not work well according to the preliminary analysis. We do not provide a full comparison with our main solution, but we find it useful to share them because this is the first generation challenge of feedback comment generation and there seems to be no or very little relevant work. 

First, we tried various ways to enrich the training samples with auxiliary information, such as part of speech of the words inside the error span, corrected words, or grammar error classes (obtained with ERRANT\footnote{\url{https://pypi.org/project/errant/}} classification model). The results of this group of approaches were slightly worse than the main solution, however, we may assume that we did not dedicate much time to that, so it could be promising to conduct further experiments in this direction.

Second, we tried to find some easy heuristics that can be used for retrieving the existing suitable comment from the train data.  We clustered learners' texts using similar words or parts of speech within the error span or vector representation of the text and then manually analyzed the feedback comments corresponding to every cluster. We also tried to do similar experiments in the opposite direction (clustered feedback comments and analyze the learners' sentences). This approach let us find some heuristics that were used as an auxiliary tool for the language model-based feedback generation. However, the decision that used this tool with the trained model did not show any significant improvement over the pure model-based approach, which most probably means that such heuristics can be learned by the language model itself.

\section{Results}

The results of the system output are scored automatically and manually. 

Automatic and manual scoring compares the system's outputs with manually created feedback comments. The automatic approach uses the BLEU score. In terms of manual scoring, a system output is regarded as appropriate if it contains information similar to the reference and does not contain information that is irrelevant to the offset; it may contain information that the reference does not contain as long as it is relevant to the offset. If these conditions are met, the output is labeled as correct. The task definition (see Section~\ref{sec:task_definition}) allows systems to generate <NO\_COMMENT> phrase which is excluded from both the numerator and the denominator of precision and the numerator of recall. That is why the final score is calculated as precision, recall, and F1-score. 

We do not make any filtering of the generated feedback, thus there is no case when our system generates <NO\_COMMENT> phrase, so all metrics are equal. Refer to \tablename~\ref{tab:results} for the results of manual evaluation by organizers of top-3 solutions. Our solution took second place.

It is also worth mentioning that the approach of using double brackets as a signal of the exact location of the error span to train the model worked well. To be more precise, the system always generates different feedback for similar sentences with different error spans and 12 out of 20 sentences in the test set (similar sentences with different slots were presented by pairs, so in total there are ten pairs of such samples) were scored as correct by organizers. Examples of such sentences can be found at \tablename~\ref{tab:similar_sentences_output}.

\section{Future work}

There are several possible ways to improve the proposed data augmentation approach, that we leave for future work. 

In our approach, we use the large language model to generate a new text using the prompt that contains a grammar error. The error could generally affect the quality of the generated text, which is why it could be interesting to first automatically correct the error in the clipped sentence, use the language model to generate a new version of the sentence, and then replace the corrected word with the erroneous word in the new sentence. 

Another promising direction of the improvement of the augmentation approach is to apply changes not only to the right part of the error span but also to the left part. This could be done, for example by filling the masks placed on the position of random words that are not syntactically related to the words within the error span.

The amount of data generated for our experiments was based on a "the good the better" basis. However, it is also worth studying the relation between the amount of augmented data and the improvement in the quality of the trained model.

\begin{table}[t!]
\begin{tabular}{lllll}
\hline
\multicolumn{1}{c}{\textbf{\#}} & \multicolumn{1}{c}{\textbf{Team ID}} & \multicolumn{1}{c}{\textbf{Precision}} & \multicolumn{1}{c}{\textbf{Recall}} & \multicolumn{1}{c}{\textbf{F1}} \\ \hline
1 & ihmana & 0.6244 & 0.6186 & 0.6215 \\
2 & nigula (ours) & 0.6093 & 0.6093 & 0.6093 \\
3 & TMUUED & 0.6132 & 0.6047 & 0.6089 \\ \hline
\end{tabular}
\caption{\textbf{Results.} Manual evaluation by organizers.}
\label{tab:results}
\end{table}

\section{Conclusion}

In this paper, we present our solution for GenChal 2022 shared task dedicated to feedback comments generation to improve the English language learning experience. Our solution uses the error span based preprocessing of the learner's text, augmentation of the dataset by clipping of the learner's text w.r.t syntactic dependency to the words within the error span, and then the inference of large language model, using clipped text as a prompt, and finally training large T5-based model with both initial and augmented version of the dataset. Our solution took second place in this competition according to manual evaluation by organizers. The model and code of our experiments are open-sourced.

We also share the track of unsuccessful experiments and general ideas about alternative approaches to this task to prepare the ground for future researchers.

\section{Acknowledements} 

This work was supported by a joint MTS-Skoltech laboratory on AI.

\bibliography{acl_latex}

\appendix



\end{document}